# Application of the Modified 2-opt and Jumping Gene Operators in Multi-Objective Genetic Algorithm to solve MOTSP

Rohan Agrawal
Department of Computer Science and IT
Jaypee Institute of Information and Technology
Noida, India
rohan.agrawal.89@jiitu.org

*Abstract*— **Evolutionary Multi-Objective Optimization is becoming a hot research area and quite a few papers regarding these algorithms have been published. However the role of local search techniques has not been expanded adequately. This paper studies the role of a local search technique called 2-opt for the Multi-Objective Travelling Salesman Problem (MOTSP). A new mutation operator called Jumping Gene (JG) is also used. Since 2-opt operator was intended for the single objective TSP, its domain has been expanded to MOTSP in this paper. This new technique is applied to the list of KroAB100 cities.**

## I. INTRODUCTION

Multi-objective traveling salesman problem (MOTSP) is a typical multi-objective optimization problem. It requires the selection of an optimal route while maintaining a balance between cost assignment and distance assignment. This paper shows a simple model using Elitist Non-dominated Sorting Genetic Algorithm (NSGA-II) with a modified 2-Opt and Jumping Gene adaptation to solve the MOTSP. The algorithm use integer coding method or real variables, creation of an initial population that satisfies the rules of a valid tour; calculation of the two objectives: distance and cost, and then ranks the chromosomes with Pareto function and uses tournament selection to select the better chromosomes to form a series of parents, through crossover, mutation and local search. The method of local search used is modified 2-opt. The results of real examples of MOTSP such as the KroAB100 list of cities demonstrate that the approximate global optimal solution set of the problem can be quickly obtained and with high accuracy.

## II. PROBLEM STATEMENT

Given a number of cities and the cost of travel between each pair of them, the travelling salesman problem is to find the cheapest tour of visiting each city exactly once and returning to the starting point. The single objective TSP is NP-hard. Mathematically, in the multi-objective TSP, the decision space is the set of all the permutations of (1, 2, ... , n), and the m objectives to minimize are:

$$f_1(x_1,...x_n) = \sum_{i=1}^{n-1} d^1_{x_i,x_{i+1}} + d^1_{x_n,x_1} \qquad (1)$$

....
....

$$f_m(x_1,...x_n) = \sum_{i=1}^{n-1} d^m_{x_i,x_{i+1}} + d^m_{x_n,x_1} \qquad (2)$$

In (1) and (2) stated above, $x = (x_1; : : : ; x_n)$ is a permutation vector, and $d^k_{i;j}$ can be regarded as the travel cost from city $i$ to $j$ in the $k^{th}$ objective, I have taken $d^k_{i;j}$ to be the Euclidean distance between $x_i$ and $x_j$ for the $k^{th}$ objective.

The data sets I have used are KroA100 and KroB100 which are 100 city single objective problems but when combined to form KroAB100 can be used as a bi-objective problem.

## III. ALGORITHM DESCRIPTION

Initially a random population $P_o$ is created. The population is sorted based on non-domination. Each solution is assigned fitness, rank based on its non-domination level, where 1 is the best level, 2 is the next based and so on. In the NSGA-II algorithm, fitness minimization is assumed. At first, the usual binary tournament selection, recombination and mutation operators (JG) are used to create an offspring population $Q_o$ of size N. But in this implementation, instead of simple mutation, I have used the Jumping Gene (JG) operator along with the modified 2-opt operator. After the initial generation, the procedure is different. The procedure for the $i^{th}$ generation of the algorithm is described in Fig. 1 below.

A combined population of $P_i$ and $Q_i$ is formed. $P_i$ is the population formed by the previous generation. $Q_i$ is the population formed after performing the various operators on the population. The combined population size is double that of the allowed population size, N. The combined population is sorted according to non-domination. This method insures elitism by making sure that the fittest individuals of the old population are present in the new one. Out of this sorted combination, we pick the best N individuals.

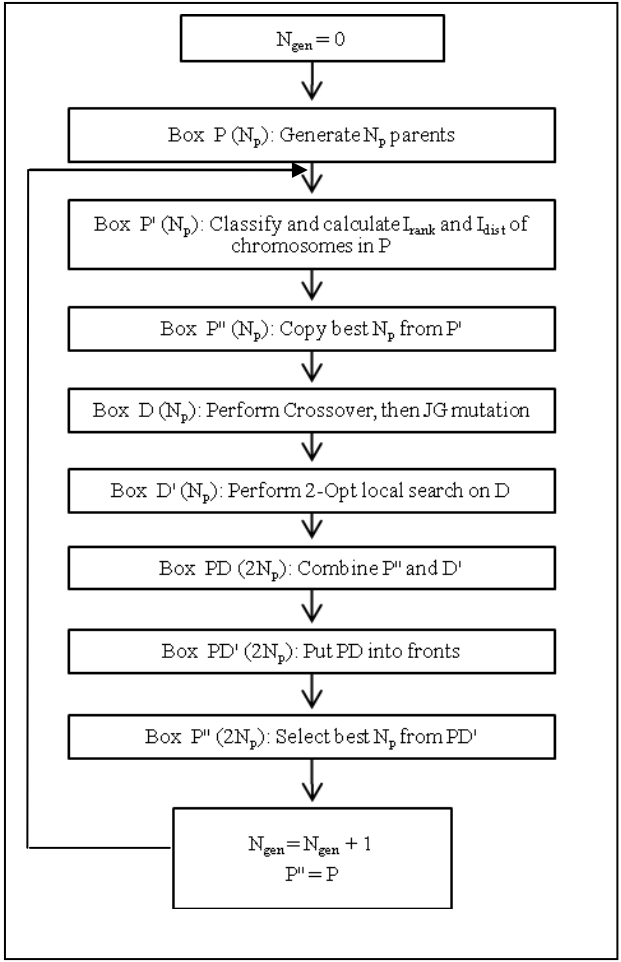

Figure 1. Flowchart of the Algorithm NSGA-II + JG + 2-Opt. A Box denotes an entire population.

In a particular non-dominated set, individuals are arranged in the decreasing order of crowding distance. Reference [1] can be looked into for a more detailed description. In Fig. 1, $N_{gen}$ is the number of generations. $N_p$ is the number of individuals in the population. A box represents an entire generation.

*A. Chromosome Encoding*

The chromosome consists of an array of cities $<x_1; \ldots ; x_n>$ and this representation means that city $x_1$ is connected with $x_2$, $x_2$ with $x_3$ and so on. Finally $x_n$ is connected to $x_1$ to form a tour. A check has to be performed while initialization to make sure that no city appears twice in the tour.

*B. Jumping Gene Operator*

Each chromosome is first checked to see if the operator will be carried out at all or not using a probability, $P_{jg}$. If the jumping gene operator is to be applied, two locations are identified randomly on the individual, and the binary sub-string in-between these points is replaced with a newly, randomly generated string of the same length. In our case, instead of binary variables, we have used real variables; therefore the value of each variable between positions p and q will be reinitialized randomly. This operator is different from normal mutation in the sense that it changes whole segments of an individual, instead of just a single chromosome. For looking into the performance advantage of JG operator over simple mutation, the reader can refer [2]. Fig. 2 shows the steps involved in the JG operator.

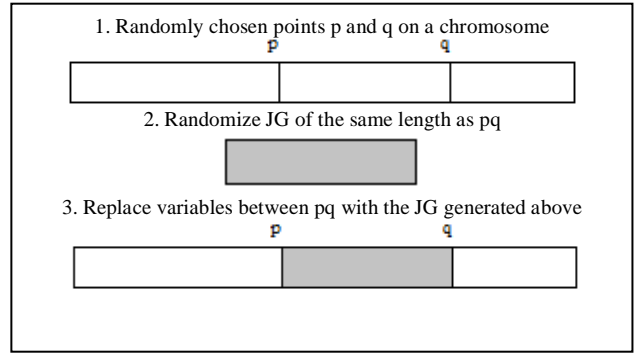

Figure 2. Steps involved in The JG operator.

*C. 2-OPT Operator*

2-OPT is a simple local search/improvement algorithm for TSP based on simple tour modifications. Given a feasible tour, the algorithm repeatedly performs operations, so long as each operation reduces the length of the current tour, until a tour is reached for which no operation yields an improvement. This tour is a locally optimal tour. 2-Opt was first proposed in [3] for single objective travelling salesman problem. The move deletes two edges, thus breaking the tour into paths, and then reconnects those paths in another possible way.

Let us consider the case of 2-Opt for single objective TSP, and let us assume a fixed orientation of the tour, with each tour edge having a unique representation as a pair (x, y) where x is the immediate predecessor of y in tour order. Then each possible 2-Opt move can be viewed as corresponding to a 4-tuple of cities < t1, t2, t4, t3> where (t1, t2) and (t4, t3) are the oriented tour edges deleted and (t2, t3) and (t1, t4) are the edges that replace them. For an improving move it must be the case that either d(t1, t2) > d(t2, t3) or d(t3, t4) > d(t4, t1) or both. As shown in Fig. 3, the newly formed tour having 4-tuple <t1, t4, t3, t2> has shorter distance. Here d(x, y) is used to represent the distance between cities x and y which is to be minimized.

*D. Modified 2-OPT Operator*

The 2-Opt operator will have to be slightly modified for use with Multi-Objective TSP. A simple way to apply the 2-Opt operator would be to randomly apply it either to the first objective, or the second objective, and so on, with equal probabilities. Let's call this method A. But this yields an uneven non dominated set, with a majority of the points lying at the extremities.

An alternative would be to sum up the objective function and use d(x, y) as the sum of the individual costs or distances between the cities x and y. Let's call this method B. This method results in a dense mid-section in the non-dominated set, but values do not reach extreme values of the objectives.

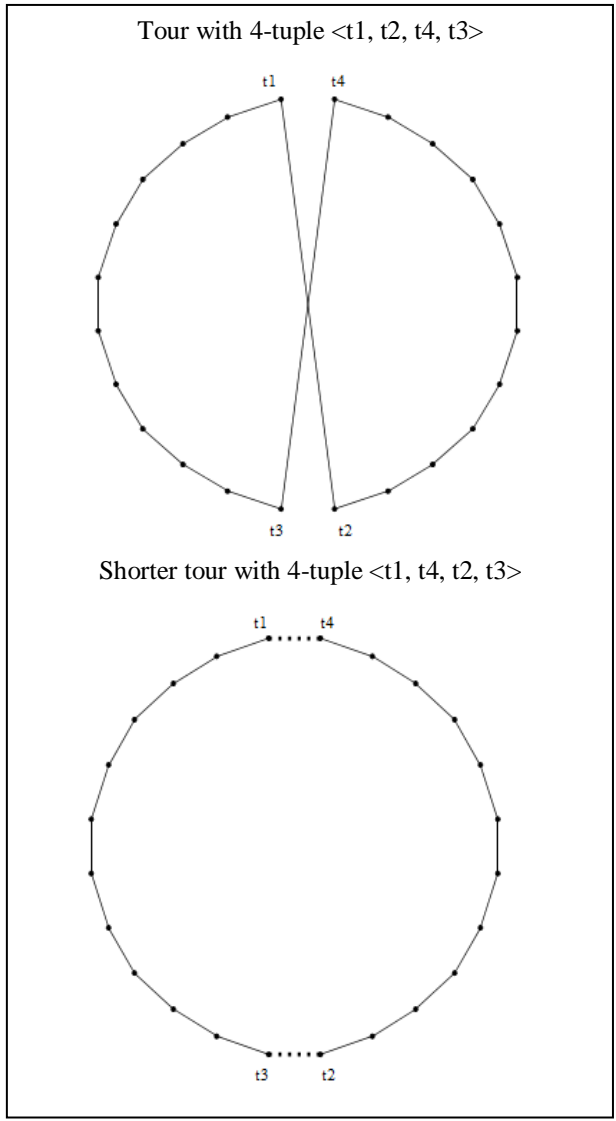

Figure 3. Tour with 4 tuple <t1,t2,t4,t3> being replaced by 4 tuple <t1,t4,t2,t3> as a result of the 2-Opt operator. The second tour is visibly shorter than the first tour.

Thus we combine these two measures for the optimal result. The probability of carrying out method B is fixed at 50%. The probability of carrying out method A therefore would be 50%.

But the probabilities of carrying out the 2-opt operator either with objective A or objective B would be further divided to 50% each. Thus this method focuses on the extremities as well as the mid-section of the non-dominated set.

## IV. RESULTS

To test our method, we have used the KroAB100 test data set which is a combination of KroA100 and KroB100 data sets. The parameters used for the GA were, population = 400, generations = 5000, Probability of crossover = 0.9, Probability of Jumping Gene = 0.5. As you can see in Fig. 4, when the 2-Opt operator is not modified for MOTSP, the mid-section of the non-dominated set is very sparse. But when the modified 2-Opt operator is used as in Fig. 5, the non-dominated set has a very even spacing throughout reaching extreme values of the individual objectives.

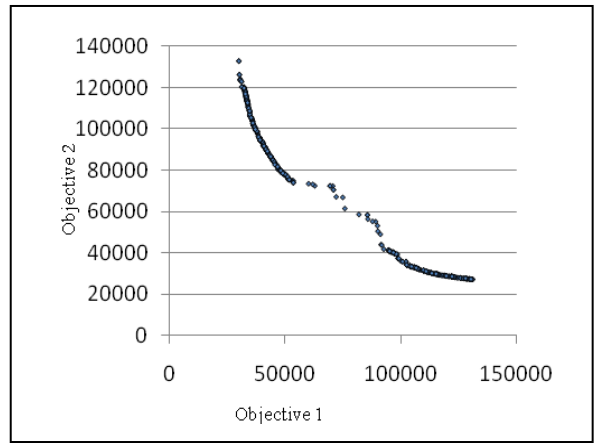

Figure 4. Non-Dominated set of solutions with KroA100 fitness values on y-axis and KroB100 fitness values on the x-axis. The unmodified 2-Opt operator is used. The mid-section of the solution set is unfavourably sparse.

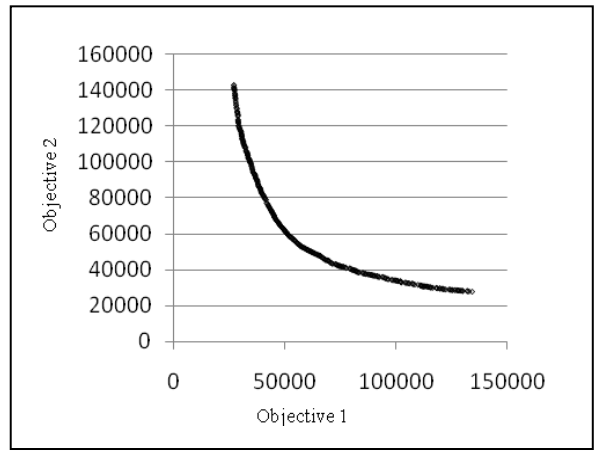

Figure 5. Non-Dominated set of solutions with KroA100 fitness values on y-axis and KroB100 fitness values on the x-axis. Modified 2-Opt operator is used. The points in the solution set are favourably even spaced.

ACKNOWLEDGMENT

I would like to thank Dr S K. Gupta and Pranava Chaudhary, both of whom have profoundly helped me in understanding Genetic Algorithm and Multi-Objective Optimization.